\documentclass[conference]{IEEEtran}

\usepackage{amsmath,amssymb, graphicx}
\usepackage[ruled,linesnumbered]{algorithm2e}
\usepackage{algorithmic}
\usepackage{graphicx}
\usepackage{paralist} 
\usepackage{color}
\usepackage{multirow}
\usepackage{rotating}
\usepackage{graphicx,wrapfig,lipsum}
\usepackage{hyperref}
\usepackage[mathscr]{eucal} 
\usepackage{amsbsy} 
\usepackage{subfigure}
\usepackage{booktabs}
\usepackage{xcolor}
\usepackage{tikz}
\usetikzlibrary{snakes}
\usepackage[outercaption]{sidecap}  
\usepackage{multirow}
\usepackage{color}
\usepackage{epstopdf}
\usepackage{balance}
\usepackage{tablefootnote}
\usepackage{empheq} 
\usepackage{moresize}
\usepackage{graphicx}

\usepackage{enumitem}
\setlist{nolistsep}
\usepackage{mdframed}
\usepackage{stackengine} 
\usepackage{sidecap}

\usepackage[font=small,labelfont=bf,skip=0pt]{caption}

\usepackage{url}

\newcounter{ALC@tempcntr}
\newcommand{\mourmeth}{\text{PAE}}
\newcommand{\newmethod}{$\mourmeth$\xspace}

\newcommand{\hide}[1]{}

\newcommand{\bit}{\begin{compactitem}}
\newcommand{\eit}{\end{compactitem}}
\newcommand{\ben}{\begin{compactenum}}
\newcommand{\een}{\end{compactenum}}

\def\x{{\mathbf x}}

\begin{document}
 
\title{\newmethod: LLM-based Product Attribute Extraction for E-Commerce Fashion Trends}

\author{

\IEEEauthorblockN{Apurva Sinha}
\IEEEauthorblockA{\textit{Walmart Global Technology} \\
apurvasinha2003@gmail.com}
\and
\IEEEauthorblockN{Ekta Gujral}
\IEEEauthorblockA{\textit{Walmart Global Technology} \\
egujr001@ucr.edu}
}
\maketitle

\begin{abstract}
Product attribute extraction is an growing field in e-commerce business, with several applications including product ranking, product recommendation, future assortment planning and improving online shopping customer experiences. Understanding the customer needs is critical part of online business, specifically fashion products. Retailers uses assortment planning to determine the mix of products to offer in each store and channel, stay responsive to market dynamics and to manage inventory and catalogs. The goal is to offer the right styles, in the right sizes and colors, through the right channels. When shoppers find products that meet their needs and desires, they are more likely to return for future purchases, fostering customer loyalty. Product attributes are a key factor in assortment planning. In this paper we present \newmethod, a product attribute extraction algorithm for future trend reports consisting text and images in PDF format. Most existing methods focus on attribute extraction from titles or product descriptions or utilize visual information from existing product images. Compared to the prior works, our work focuses on attribute extraction from PDF files where upcoming fashion trends are explained.  This work proposes a more comprehensive framework that fully utilizes the different modalities for attribute extraction and help retailers to plan the assortment in advance. Our contributions are three-fold: (a) We develop \newmethod, an efficient framework to extract attributes from unstructured data (text and images); (b) We provide catalog matching methodology based on BERT representations to discover the existing attributes using upcoming attribute values; (c) We conduct extensive experiments with several baselines and show that \newmethod is an effective, flexible and  on par or superior (avg $92.5\%$ F1-Score) framework to existing state-of-the-art for attribute value extraction task.
\end{abstract}
\section{Introduction}
\label{sec:intro}
Assortment planning for future products plays a crucial role in the success of e-Commerce as a platform. It involves strategically selecting and organizing a range of products to meet customer demands and maximize sales. This process involves analyzing market trends, customer preferences, and competitor strategies to identify potential gaps and opportunities. By carefully planning the assortment, retailers can ensure they offer a diverse and relevant range of products that cater to different customer segments. This helps in driving customer satisfaction, increasing sales, and staying ahead in the competitive market. Walmart collaborate with trend forecasting company that provides insights and analytics for the fashion and creative industries. They do not release public reports, as their insights are provided through a paid subscription service. However, they often share snippets of their forecasts via blog posts or on social media. For example, they might report on upcoming color trends for a particular season, predict consumer behaviors, or identify emerging fashion trends in different regions. The trend forecasting company also provides reports on retail and marketing strategies, textiles and materials innovations, product development and lifestyle and interiors trends. Their reports are typically used by retailers and marketers to plan and develop their products and strategies.

\textit{Informal Problem 1. Given a set of target attributes (e.g., color, age group, material), and unstructured information in the form of  text and images: how can we extract values for the attributes? What if some of these attributes have multiple values, like colors or age group?}

Correct predicted attributes helps in improved catalog mapping, which helps in generating search tags on better content quality of products. Customers can filter for products based on their exact needs and compare product variants promptly. Resulting in a seamless shopping experience while searching or browsing a product on an E-commerce platform. The Product attribute Extraction (PAE) engine can help the retail industry to onboard new items or extract attributes from existing catalog.
\begin{figure}
	\begin{center}
	    \includegraphics[width=0.45\textwidth]{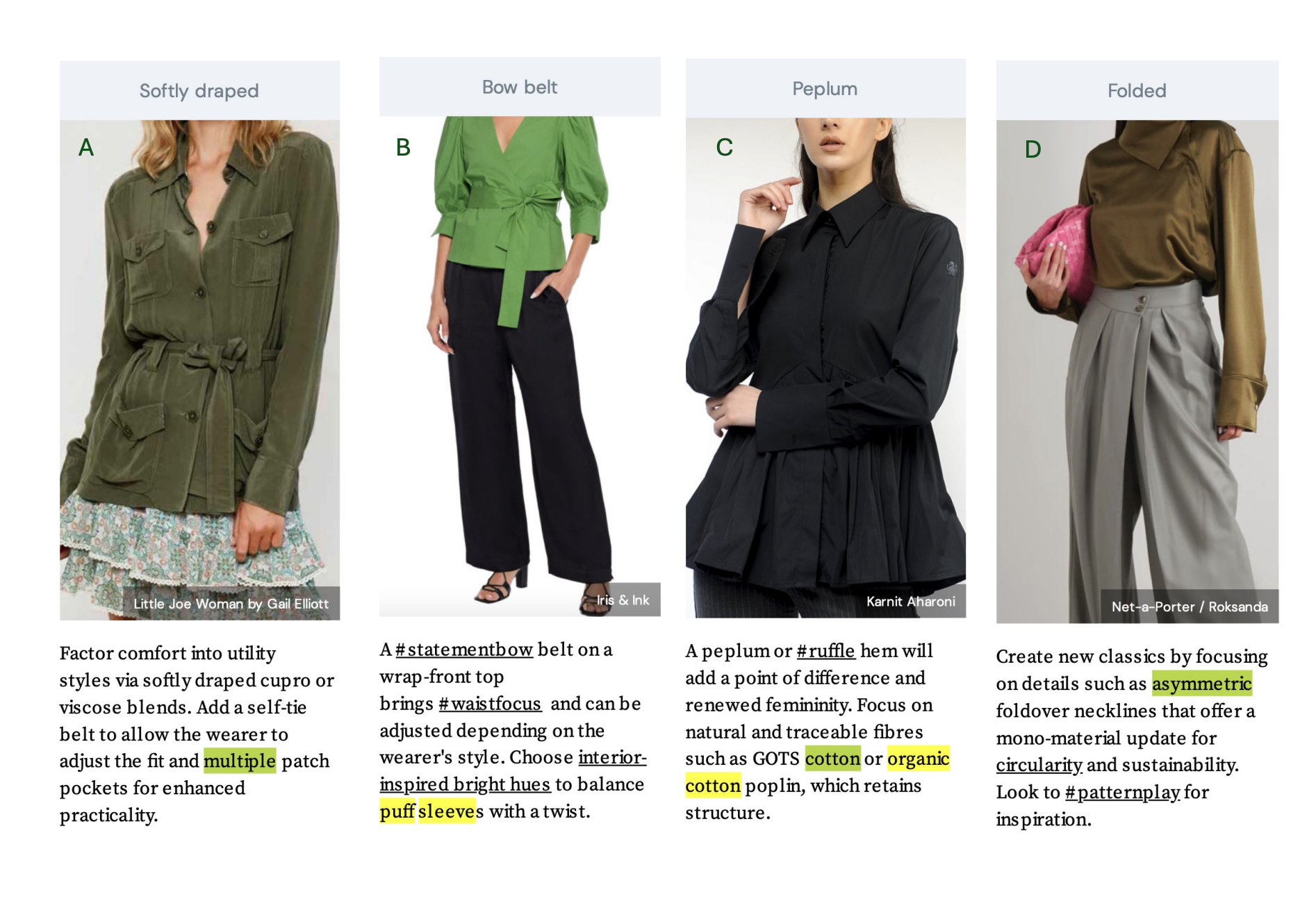}
		\caption{Example of Text and Images for Attribute Extraction} 
		\label{fig:exampleproblem}
	\end{center}
\end{figure}

\begin{figure*}
	\begin{center}
	    \includegraphics[clip,trim=0cm 0cm 0cm 0cm,width=0.95\textwidth]{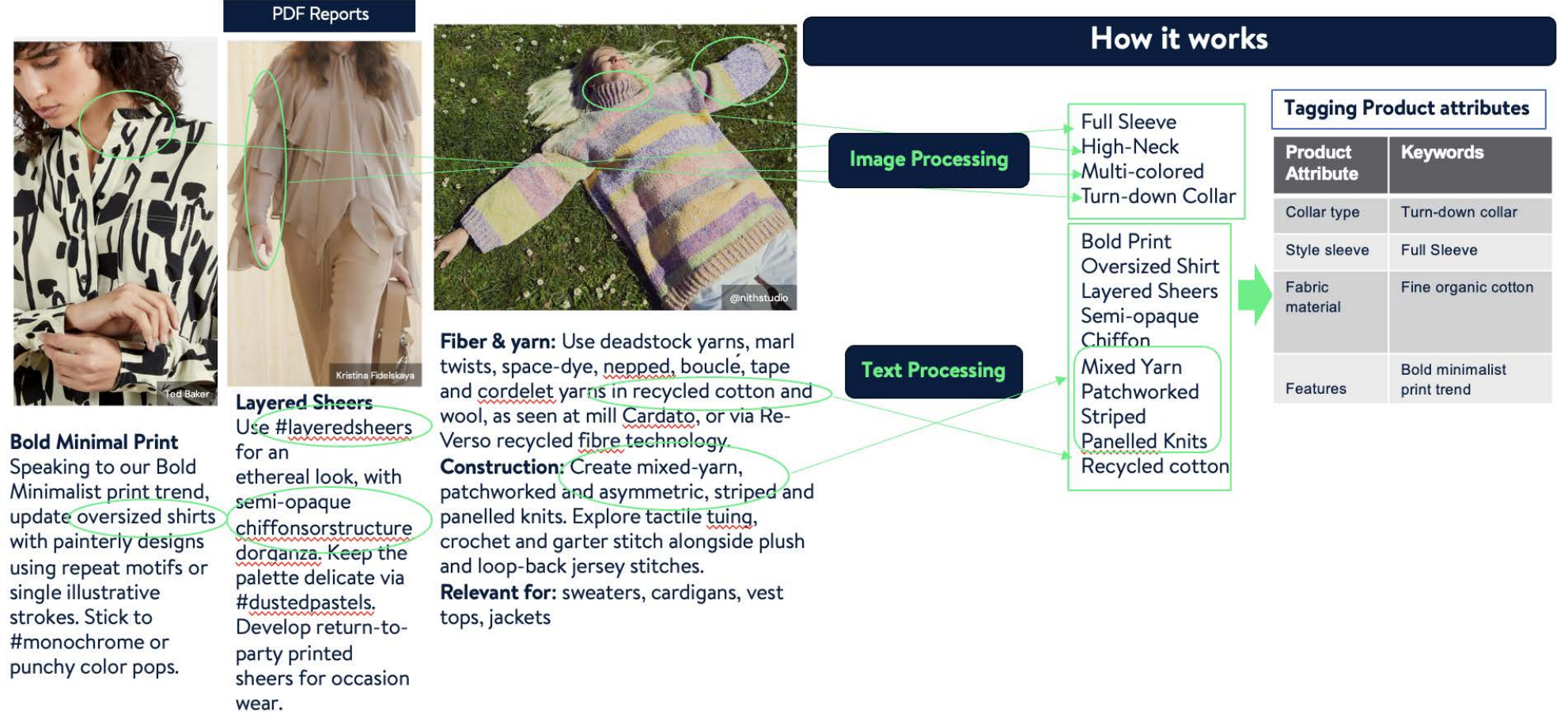}
		\caption{End to End PAE mapping to product catalog} 
		\label{fig:end_to_end_flow}
	\end{center}
\end{figure*}

\textbf{Motivating Example}
Retailers can use upcoming market trends to decide on product catalog assortment planning based on upcoming seasons like spring, fall and summer. For a concrete example, refer to Figure \ref{fig:exampleproblem}, a classic shirt (with unstructured text and image data), it talks about \textit{peplum} or \textit{ruffle hem} being used on Global Organic Textile Standard \textit{(GOTS) cotton} or \textit{organic cotton poplin}. This can be referenced to having classic shirts in a catalog made of Organic/GOTS cotton and peplum/ruffle hem as features for the shirt. Based on these attribute insights, assortment planners would work closely with suppliers and designers to curate a collection of such clothing items to complete the look, including leggings, sports bras, sweatshirts, and sneakers. They would consider factors such as quality, affordability, and inclusively to ensure that the assortment caters to a wide range of customers. The images shows popular prints, innovative fabrics, and style variations within the classic shirt category. This would enable retailers to offer a diverse selection of classic shirt options that align with the latest trends. Additionally, once retailers have an assortment ready for selling, they could collaborate with fashion influencers who align with the upcoming trend to create exclusive collections or promote the existing assortment. This would help to generate excitement among customers and drive high engagement. By incorporating the recommended color palettes, visual elements, and messaging, the retailer could create an immersive shopping experience.

\textbf{Previous Works} 
In this section, we provide a brief overview of existing Multi Modal Attribute Extraction (MMAE) techniques being used to extract product attributes from Images and Text. MMAE explained in paper \cite{logan2017multimodal} talks about returning values for attributes which occur in images as well as text and they do not treat the problem as a labeling problem. They define the problem as following: Given a product $i$ and a query attribute $a$, extract a corresponding value $v$ from the evidence provided in terms of textual description of it ($D_i$) and a collection of images ($I_i$). For training, for a set of product items $I$, for each item $i \in I$, its textual description Di and the images $I_i$, and a set $A_i$ comprised of attribute-value pairs $A_i = \{ \langle a^j_i, v^j_i \rangle \}_j$. The model is composed of three separate modules: (1) an encoding module that uses modern neural architectures to jointly embed the query, text, and images into a common latent space, (2) a fusion module that combines these embedded vectors using an attribute-specific attention mechanism to a single dense vector, and (3) a similarity-based value decoder which produces the final value prediction.  Another approach for MMAE explained in paper \cite{de2022multi}, considers cross-modality comparisons. They leverage pre-trained deep architectures to predict attributes from text or image data. By applying several refinements to leverage pre-trained architectures and build  single modality models like Text only modality model, image only modality model for the task of product attribute prediction. A new modality merging method was proposed to mitigate modality collapse. For every product, it lets the model assign different weights to each modality and introduces a principled regularization scheme to mitigate modality collapse. Paper by \cite{zhu2020multimodal} talks about Multi modal Joint Attribute Prediction and Value Extraction for E-commerce Product. They enhance the semantic representation of the textual product descriptions with a global gated cross-modality attention module that is anticipated to benefit attribute prediction tasks with visually grounded semantics. Moreover, for different values, the model selectively utilizes visual information with a regional-gated cross-modality attention module to improve the accuracy of value extraction. Note: As these methods are industry related, hence source code is not publicly available to reproduce the outcome.

\textbf{Challenges} 
Despite the potential, leveraging PDF reports consisting text and images for attribute value extraction remains a difficult problem. We highlight few challenges faced during designing and executing  extraction framework:
\begin{itemize}
    \item \textbf{C1: Text Extraction from PDF}: PDF reports can be a combination of multiple images, overlapping text elements, annotations, metadata and unstructured text integrated together in no specific PDF format. Extracting text from such reports can be difficult, challenging and lead to misspelled text and loss of specific topic-related context. Another issue is missing and noisy attributes. Text data might not have all the attributes which we are looking for. Therefore, visual attribute extraction plays an important role.
    \item \textbf{C2: Image Extraction from PDF}: Images in PDF reports can be embedded, compressed down to reduce size, in various formats like JPEG, PNG etc. Extracting images while maintaining the resolution and quality of images requires specialized handling to accurately preserve the original appearance. Also, images could bring multi-labeled attributes which can confuse the model but can be mitigated by merging certain attribute values to help with model inferences.  
    \item \textbf{C3: Extracting Product Attributes}: Product tags extracted from text/images needs to be carefully mined to match product attributes. The attributes differ based on the category of products we are referring to and can have multi-labeled attributes. For example, women's tops will have sleeve related attribute whereas women's trousers will have type of fit attribute and sleeve attribute will be irrelevant.
    \item \textbf{C4: Mapping Product Attributes to Product Catalog}: E-commerce catalog has specific products and attributes mapped to them. On-boarding new attributes based on PDF reports, requires new attribute creation/refactoring existing attributes.
\end{itemize} 

\textit{Informal Problem 2. Can we develop unsupervised models that require limited human annotation? Additionally, can we develop models that can extract explainable visual attributes, unlike black-box methods that are difficult to debug?}

\textbf{Mitigating Challenges:} 
Current multi-modal attribute extraction solutions \cite{logan2017multimodal,de2022multi} are inadequate in the e-commerce field when it comes to handling challenges C2 and C4. Conversely, text extraction solutions that successfully extract attribute values are primarily text-oriented \cite{GHOSH2023,zheng2018opentag,bougouin2013topicrank,xu2019scaling} and cannot be easily applied to extracting attributes from images. In this work, we address the central question: how can we perform multi-modal product attribute extraction from upcoming trend PDF reports? The detail description is given in section \ref{sec:method} to handle each challenge. Our proposed method \newmethod works on extracting upcoming trends from PDF reports generated by the trend forecasting company. This capability provides an insight into upcoming marketing trends and customer preferences. By using trend forecasting reports, catalog can be refined with new classes of products having trending attributes based on external reports, to propel value across the apparel space by accurately indicating attribute trends in the market and increasing customer satisfaction. The contributions of our paper are as follows:
\begin{itemize}
\item \textbf{Novel Problem Formulation}: We propose the end-to-end model of jointly extracting the trending product attributes and hashtags from PDF files consisting of text and image data and mapping it back with the product catalog for the final product attributes values. An example of end to end execution of product attribute extraction and mapping is shown in the figure \ref{fig:end_to_end_flow}. Due to Walmart Privacy Requirements, models and datasets are not open to public. We have elaborated the details of each model, and readers can use LLM model of their own choice.

\item \textbf{Flexible Framework}: We develop a general framework \newmethod for extracting text and images from PDF files and then generating product attributes. All the components are easily modified to enhance the capability or to use the framework partially for other applications. The extraction engine can be used to extract attributes for different categories of products like Electronics, Home decor etc.

\item \textbf{Experiments}: We performed extensive experiments in real-life datasets to demonstrate \newmethod’s efficacy. It successfully discovers attribute values from text and image data  with a high F1-score of 96.8\%, outperforming state-of-the-art models. This proves its ability to produce stable and promising results.
\end{itemize}
The remainder of the paper is organized as follows: The problem formulation is given
in Section \ref{sec:problem}. In Section \ref{sec:method} we describe our proposed method \newmethod in details with examples. Finally, we show the experimental results in section  \ref{obtd:experiments} and section  \ref{sec:conclusions} concludes the paper.
\begin{figure*}
	\begin{center}
	    \includegraphics[clip,trim=0cm 2cm 0cm 2cm,width=0.95\textwidth]{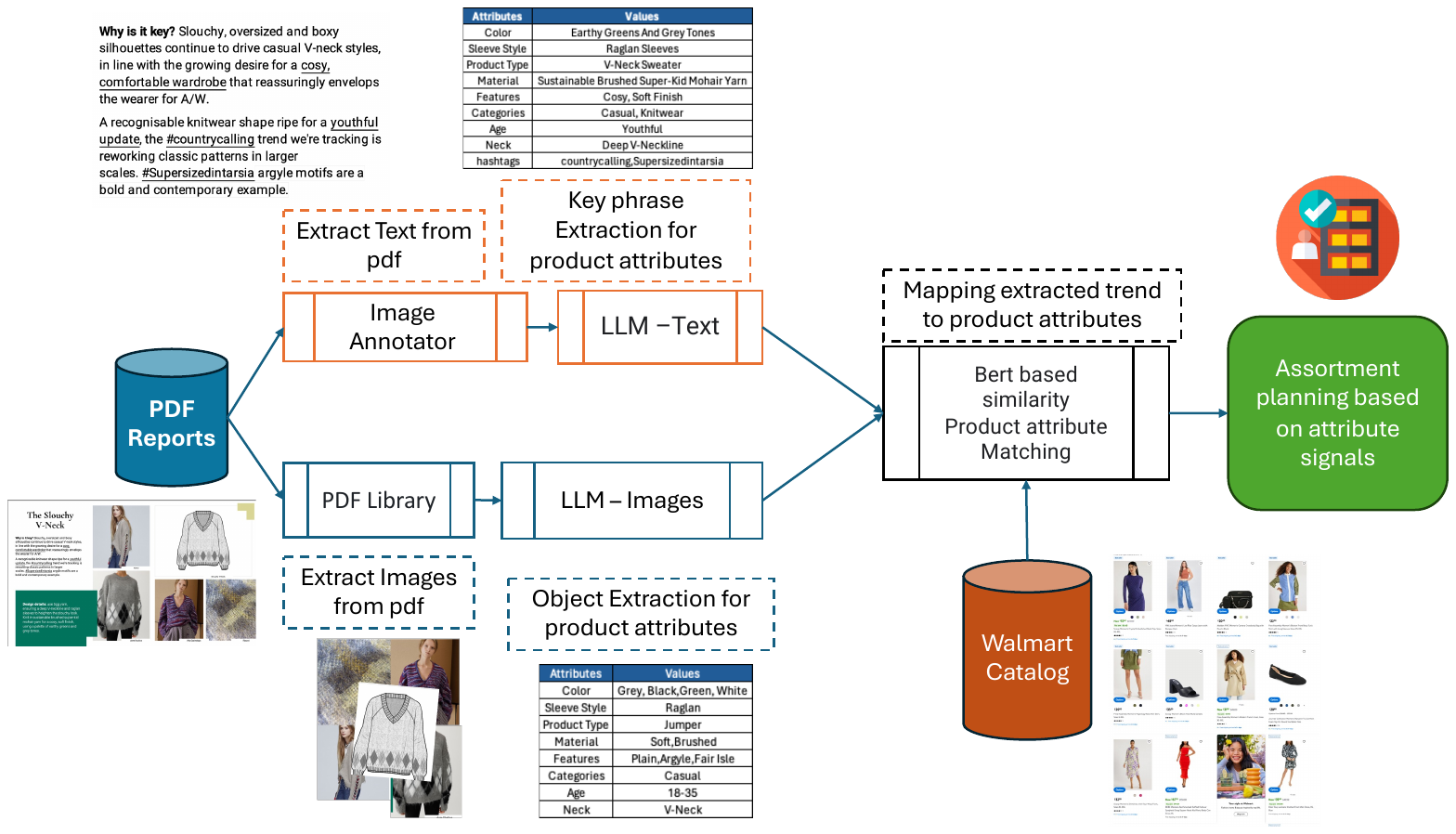}
		\caption{\bf Overview of the proposed Product Attribute Extraction Engine. 1) PDF Data Extraction: Text and Images. 2) Text and Image Attribute Extraction:  By using Large Language Models(LLM) models we extract various attributes like color, age etc. 3) Catalog Matching: We create attribute embeddings for extracted attributes and existing catalog attributes. Based on high cosine similarity, we send signal to future assortment planning teams.} 
		\label{fig:introimage}
	\end{center}
\end{figure*}

\section{Problem Definition}
\label{sec:problem}
The upcoming trend information in PDF files usually looks like figure \ref{fig:exampleproblem}. The text describes the upcoming trends and style types. The images along with it shows how the style will look on different models.  We consider each page of PDF file as one product type and that can be Woven Tops, Knitwear etc.  We formally define the problem as follows.

\textbf{Problem definition}: With the following information:
\begin{mdframed}[backgroundcolor= orange!20,linewidth=1.3pt,] 
\textbf{Given} (a) a PDF file with multiple pages $[1,2,3 \dots , N]$ consist of text $T_1^N$ and image $I_1^N$ data (b) LLM prompt $P$ with target attributes $attr \in$ [color, sleeve style, product type, material, features, categories, age and neck ]  \\
\textbf{Find} the value $vals$ for the target attribute $attr$ related for each page.
\end{mdframed}

Figure \ref{fig:exampleproblem} displayed a few such attribute values from Text and Image in the PDF file. Specifically, considering the target attribute \textit{Material} shown in figure \ref{fig:exampleproblem} - C, our objective is to extract the attribute value \textit{Cotton}. If the target attribute is \textit{Neck} in figure \ref{fig:exampleproblem} - B from image, the objective is to extract the attribute value \textit{V-Neck}. Similarly, target attribute \textit{Feature} for figure \ref{fig:exampleproblem} - (A, B, C, D), our objective is to extract the attribute value \textit{Softly Draped, Bow belt, Peplum and Folded} respectively.

\section{Product Attribute Extraction}
\label{sec:method}
In this work, we tackle the attribute value extraction as pair task, i.e., extracting the attribute values from image and text together. The input of the task is a “textual information $T$, set of images $I_1,I_2\dots I_N$ pair per PDF page, and the output is the product attributes values. Our framework is presented in figure \ref{fig:introimage}. In fact, through extensive experiments
(see Section \ref{obtd:experiments}), we show that our proposed method is not only intuitive, but achieves state-of-the-art performance compared to previously proposed methods in literature. Due to Walmart Privacy Requirements, models and datasets are not open to public. We have elaborated the details of each model, and you can use LLM model of your choice.

\subsection{Text Extraction from PDF}
Text extraction from PDF is an important process that involves the conversion of data contained in PDF files into an editable and searchable format. This procedure is crucial for activities like data analysis, content re-purposing, and detecting trends from public reports. However, it can pose certain challenges. The layout complexity of a PDF document can make the extraction process difficult. For instance, the presence of multiple columns, images, tables, and footnotes can complicate the extraction of pure text. Another challenge is the use of non-standard or custom fonts in PDFs, which can lead to inaccurate extraction results. Moreover, the presence of 'noise' such as headers, footers, HTML tags and page numbers can also interfere with the extraction process. 
\begin{figure}
	\begin{center}
	    \includegraphics[clip,trim=0cm 0cm 0cm 0cm,width=0.45\textwidth]{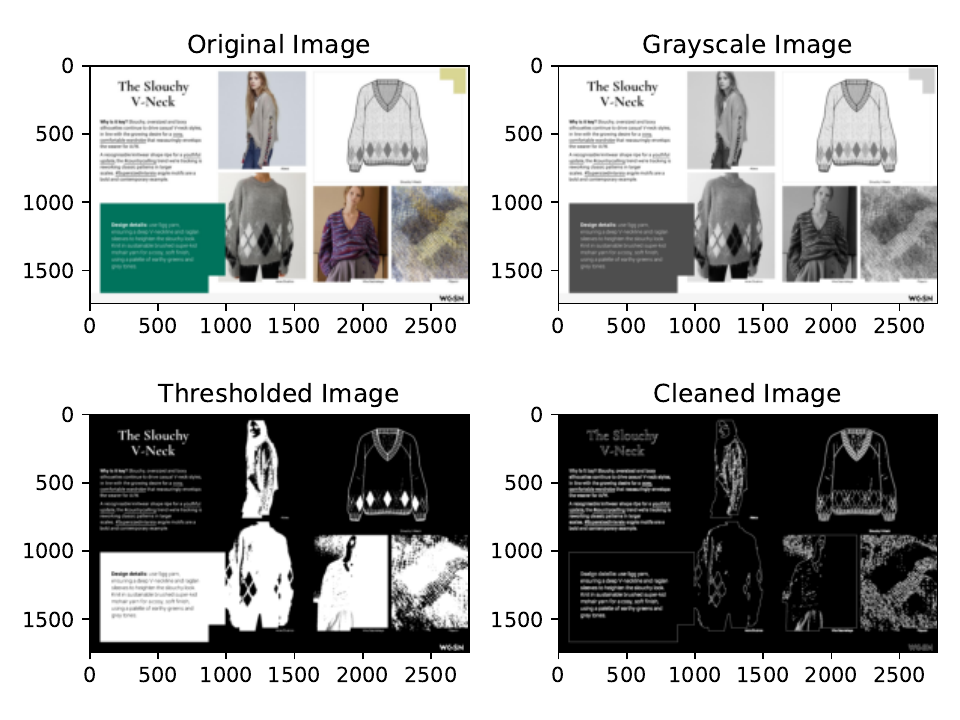}
		\caption{Text Extraction via Image Annotator} 
		\label{fig:text_extraction}
	\end{center}
\end{figure}
There are numerous tools available for text extraction from PDF files. Searching for text extraction from PDF on Google yields a plethora of results featuring various tools or pages suggesting such tools as pdfMiner \cite{pdfminer}, pdfquery\cite{pyquery} etc. However, figure \ref{fig:text_extraction} represents the process we used to extract the text from pdf files. First, we split the PDF files into PIL (Python Imaging Library) images using "convert from path" function from the pdf2image \cite{pdf2image}. Internally, the function uses the \textit{pdfinfo} command-line tool to extract metadata from the PDF file, such as the number of pages. It then uses the \textit{pdftocairo} command-line tool to convert each page of the PDF into an image. Second, we convert the images to grayscale and perform morphological transformations on each page by applying a morphological gradient operator to enhance and isolate text regions. Finally, we use Image Annotator \cite{imagegoogle} consists of Optical Character Recognition (OCR) capabilities for text extraction. Once the text is extracted, we use the spell Corrector like language-tool to fix any misinterpreted text from OCR.  
The Text extracted from PDF report about product type "Slouchy V-Neck" is given below:
\begin{mdframed}[backgroundcolor= blue!10,linewidth=1.3pt,] 
The Slouchy V-Neck Why is it key? Slouchy, over-
sized and boxy silhouettes continue to drive casual
V-neck styles, in line with the growing desire for a
cosy, comfortable wardrobe that reassuringly envelops
the wearer for A/W. A recognisable knitwear shape
ripe for a youthful update, the \#countrycalling trend
we’re tracking is reworking classic patterns in larger
scales. \#Supersizedintarsia argyle motifs are a bold and
contemporary example.
Design details: use 5gg yarn, ensuring a deep V-neckline
and raglan sleeves to heighten the slouchy look. Knit in
sustainable brushed super-kid mohair yarn for a cosy,
soft finish, using a palette of earthy greens and grey
tones.
\end{mdframed}
 
\subsection{Image Extraction from PDF}
\begin{figure}
	\begin{center}
	    \includegraphics[clip,trim=0cm 2cm 0cm 2cm,width=0.45\textwidth]{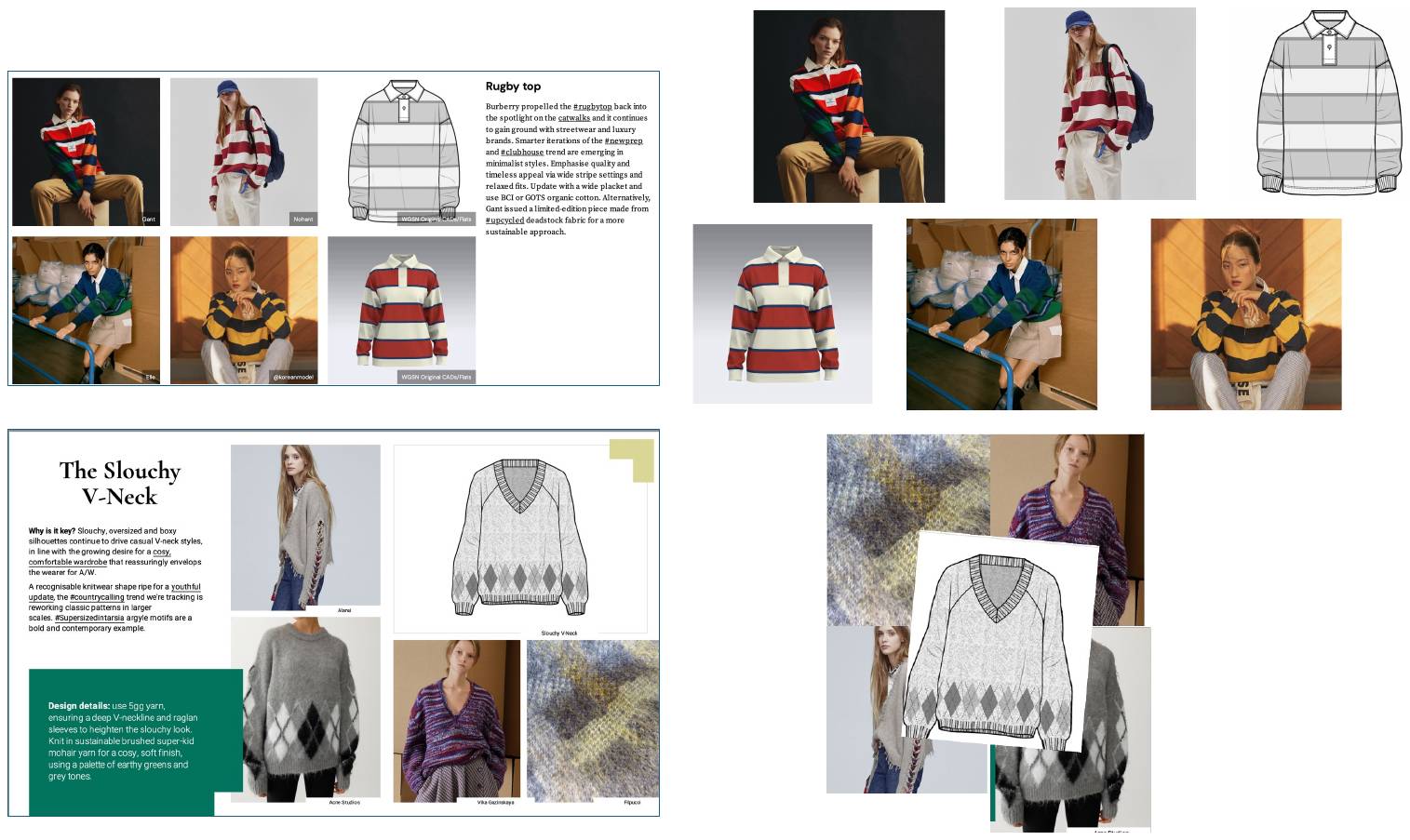}
		\caption{Example of extracted images from PDF files} 
		\label{fig:Image_extraction_from_pdf}
	\end{center}
\end{figure}

PDF files can contain images in various formats such as JPEG, PNG, or TIFF. Extracting images from different formats may require multiple techniques. The second challenge could be different types of images in the pdf files, including scanned documents, vector graphics, or embedded images.  Third, extracting images from large PDF files efficiently and in a timely manner can be a challenge, especially when dealing with limited system resources. To tackle the above-mentioned challenges, we exploit pure-python PDF library \cite{PyPDF4}, as a standalone library for directly extracting image objects from PDF files. With pure-python PDF library, we identify the pages with images and extract them as raw byte strings. Then, using Pillow, the extracted images are processed and saved in jpg formats. Figure \ref{fig:Image_extraction_from_pdf} shows extracted images from files. 
\subsection{Attribute Extraction from Text}
PDF reports consist of specific products or product categories, providing details on their design, features, materials, colors, and styles. These reports provide information about product innovations, market demand, and consumer preferences in upcoming months or years. Here, we extract 8 product attributes namely [Color, Sleeve Style, Product Type, Material, Features, Categories, Age, Neck]. We also extract hashtags to discover and explore content related to a specific topic or theme in catalog.

We are utilising the LLM model for extracting the attributes. Below is a sample prompt:

\begin{mdframed}[backgroundcolor= orange!20,linewidth=1.3pt,] 
\textbf{Generate} me color, sleeve style, product type, material, features, categories, age and neck attributes of below text:\\
a text $T_i$ 
\end{mdframed}
 
\begin{mdframed}[backgroundcolor= blue!10,linewidth=1.3pt,] 
\textbf{Example}\\
\textbf{Generate} me color, sleeve style, product type, material, features, categories, age and neck attributes from the following text: \\
\textit{"The Slouchy V-Neck Why is it key? Slouchy, oversized and boxy silhouettes continue to drive casual V-neck styles, in line with the growing desire for a cosy, comfortable wardrobe that reassuringly envelops the wearer for A/W. A recognisable knitwear shape ripe for a youthful update, the \#countrycalling trend we're tracking is reworking classic patterns in larger scales. \#Supersizedintarsia argyle motifs are a bold and contemporary example. \\
Design details: use 5gg yarn, ensuring a deep V-neckline and raglan sleeves to heighten the slouchy look. Knit in sustainable brushed super-kid mohair yarn for a cosy, soft finish, using a palette of earthy greens and grey tones."} 
\end{mdframed}

The output is then processed in dictionary type as follow:
\begin{mdframed}[backgroundcolor= blue!10,linewidth=1.3pt] 
\textbf{Output of above text example}\\
  {\textit{Color}: Earthy Greens And Grey Tones\\
 \textit{Sleeve Style}: Raglan Sleeves\\
 \textit{Product Type}: V-Neck Sweater\\
 \textit{Material}: Sustainable Brushed Super-Kid Mohair Yarn\\
 \textit{Features}: Cosy, Soft Finish\\
 \textit{Categories}: Casual, Knitwear\\
 \textit{Age}: Youthful\\
 \textit{Neck}: Deep V-Neckline}
\end{mdframed}

\subsection{Attribute Extraction from Images}
The extraction of detailed image attributes from fashion images has a wide range of uses in the field of e-commerce. The recognition of visual image attributes is vital for understanding fashion, improving catalogs, enhancing visual searches, and providing recommendations. In fashion images, the dimensionality can be higher due to the complexity and diversity of fashion items. For instance, a single piece of clothing can have multiple attributes for color, fabric type, style, design details, size, brand, and others. Hence, image attribute extraction has become more complex than text. However, these attributes can be extracted using various computer vision techniques, such as image segmentation, object detection, pattern recognition and deep learning algorithms. In this work, we explore the vision based LLM model. Each extracted image as shown in figure \ref{fig:image_example} is converted to base64 encoding. Base64 encoding is a method of converting binary data, such as an image, into ASCII text format. This is required as current LLM models takes text format as input. The ASCII text format example as follow:
\begin{figure}
	\begin{center}
	    \includegraphics[clip,trim=1cm 2cm 3cm 2cm,width=0.45\textwidth]{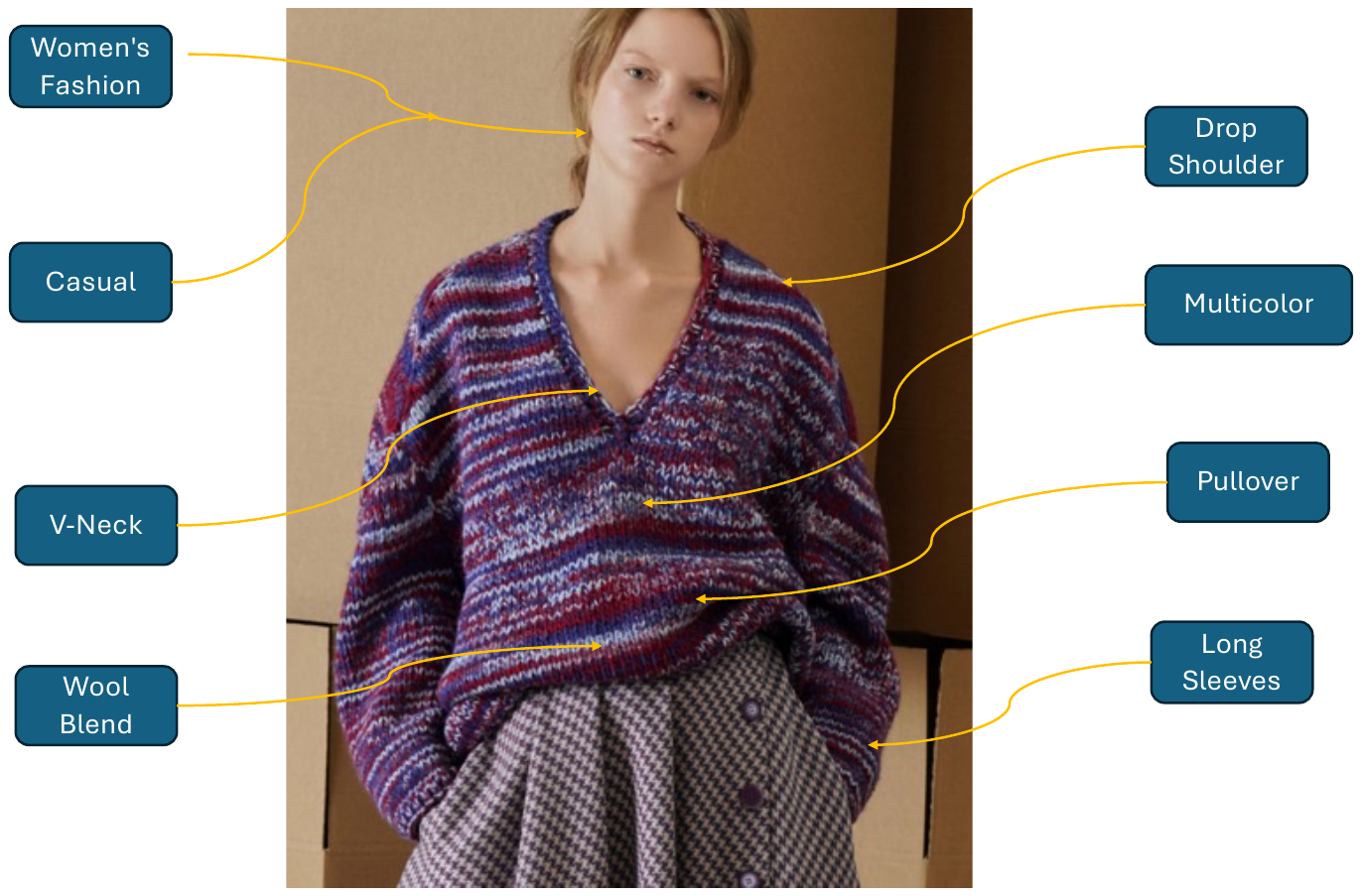}
		\caption{Image Example} 
		\label{fig:image_example}
	\end{center}
\end{figure}

\begin{mdframed}[backgroundcolor= blue!10,linewidth=1.3pt,] 
\textbf{ASCII text format}\\
\textit{'iVBORw0KGgoAAAANSUhEUgAAAU4AAA\\
GwCAIAAABJqRtXAAAACXBIWXMAAA7EAAAOxAGV\\
Kw4bAAEV4klEQVR4nOy9d0BUV9f2HRXEjgU\\
QBcWCscXeNfbebnuLHbtYsPfeY1cEG2IBbKjYF\\
 \dots'}
\end{mdframed}
Next, we use this encoded string along with LLM prompt to generate the product attributes as follow:
\begin{mdframed}[backgroundcolor= blue!10,linewidth=1.3pt,] 
\textbf{Example}\\
\textit{text: "\textbf{Generate} a list format color, sleeve style, product type, material attributes from the the below image. Also give me features, categories, age and neck attributes from below image."}\\
"inlineData": \{\\
"mimeType": "image/png",\\
"data": \textit{'iVBORw0KGgoAAAANSUhEUgAAAU4AAA\\
GwCAIAAABJqRtXAAAACXBIWXMAAA7EAAAOxAGV\\
Kw4bAAEV4klEQVR4nOy9d0BUV9f2HRXEjgU\\
QBcWCscXeNfbebnuLHbtYsPfeY1cEG2IBbKjYF\\
 \dots'}
 \}
\end{mdframed}

The output is then processed in dictionary type as follow:
\begin{mdframed}[backgroundcolor= blue!10,linewidth=1.3pt,] 
\textbf{Output of above image example}\\
  {\textit{Color}: Multicolor,\\
 \textit{Sleeve Style}: Long Sleeve\\
 \textit{Product Type}: Pullover\\
 \textit{Material}: Wool Blend\\
 \textit{Features}: V-Neck, Drop Shoulder\\
 \textit{Categories}: Women's Fashion\\
 \textit{Age}: Adult\\
 \textit{Neck}: V-Neck}
\end{mdframed}

Another common issue that arises is the presence of noisy and missing labels. It is a challenging task to accurately label and annotate all the relevant information for every page in the PDF. Despite employing various automated and manual annotation processes, it is nearly impossible to obtain perfectly labeled structured data. To address this, we employ image pre-processing or data cleaning techniques to eliminate duplicate, noisy, and invalid images before proceeding with  attribute extraction. \textit{Once we extract attributes from text and images on each page, we aggregate them per page for our further analysis.}

\subsection{Hashtag Detection in Text}
Hashtags are words or phrases preceded by the pound sign (\#) and are commonly used on social media platforms to categorize and group similar content. Detecting hashtags in text is crucial for various applications such as topic modeling, sentiment analysis, and product recommendation. The process of hashtag detection involves analyzing the text and identifying words or phrases that are preceded by the pound sign, while considering factors such as word boundaries and punctuation marks.  The extracted hashtags can then be used to gain insights into trending topics, user interests, or to enhance search and recommendation systems. In our work, we use the regular expression $'\#w+'$ to detect the hashtags.
\begin{mdframed}[backgroundcolor= blue!10,linewidth=1.3pt,] 
\textbf{Hashtags extracted from above example text}\\
\#countrycalling\\
\#Supersizedintarsia
\end{mdframed}
\subsection{Product attribute Matching}
\begin{figure}
	\begin{center}
	    \includegraphics[clip,trim=4cm 3cm 4cm 3cm,width=0.45\textwidth]{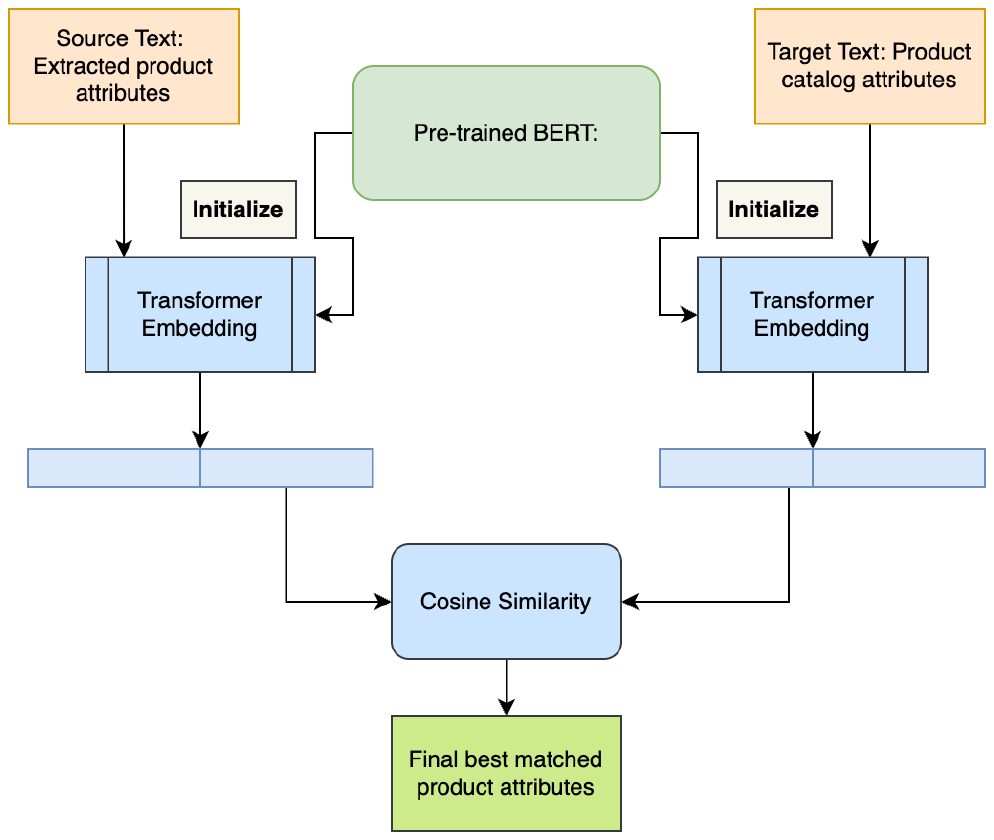}
		\caption{Product attribute Matching Framework} 
		\label{fig:bertsim}
	\end{center}
\end{figure}

The purpose of product attribute matching is to ensure that extracted attributes meet specific criteria or requirements. Based on the identified attributes, retailers plan their future product assortments. This includes selecting or designing clothing items, accessories, and other fashion products that reflect the upcoming trends. One of the challenge of product attribute matching is multiple variants of representation for the same value of one attribute. For example, "vneck" and "V-Neck" is consolidated into "V-Neck" as neck product attribute. Figure \ref{fig:bertsim} shows our framework to match the predicted attributes to the existing catalog attributes. We exploit pre-trained bert uncased model. BERT is designed to pre-train deep bidirectional representations from unlabeled text by jointly conditioning on both left and right context in all layers. We create representations or embeddings for predicted attributes and existing product attributes. Finally, we use cosine similarity to match the similar product attributes from the catalog. 

\textbf{Summary}: As shown in figure \ref{fig:end_to_end_flow}, our proposed approach relies on four successive steps. First, the \textit{text and image extraction step} that extracts all the text (paragraphs) and relevant images from the given PDF files. Second, \textit{attribute extraction step}, that uses LLM models to extract relevant attributes from the images and text. Third, \textit{merging step}, we consolidate attributes into each category and keep unique values for each attribute. Finally, \textit{catalog matching step}, helps retailers to find a product that matches with existing inventory and plan for future assortment.
\section{Experiments}
\label{obtd:experiments}
Although our work is mostly related to retail business, we will compare the performance of our  \newmethod's sub-parts with different baselines on real-life datasets. We evaluate our approach on 14 upcoming trend reports. In particular, we want to answer the following questions: 
\begin{itemize}[noitemsep]
	\item \textbf{(Q1)} How accurate is our proposed method \newmethod when compared to other baselines? 
        \item \textbf{(Q2)} How sensitive is \newmethod w.r.t different parameters?
        \item \textbf{(Q3)} How time consuming is \newmethod?
\end{itemize}
\subsection{Data-set description}
We provide the datasets used for evaluation in Table \ref{tbl:realdataset}.  These are trend reports used for 2023 assortment planning.  Assuming the attribute value is applicable to each pdf page, if the attribute value information can not be observed from the
given text description and images, we will assign “Not Mentioned” as the corresponding value. 
\begin{table}[t]
	\centering
	\footnotesize
	\begin{tabular}{|c||c|c|c|c|c|}
	\cline{1-6}
    {\bf Dataset}	& {\bf $\#P$} & {\bf $\#T$}& {\bf $\#I$} & {\bf $\#H$} &{\bf \text{\em{$GT$}}} \\	\hline
	Boys Apparel&$7$&$11$&$32$&$0$&$Y$\\\hline
         Women's Cut Sew&$7$&$30$&$24$&$5$&$Y$\\\hline
         Women's Woven Tops&$6$&$28$&$24$&$6$&$Y$\\\hline
        Country Life Boys &$12$&$12$&$66$&$3$&$Y$\\\hline
        Knitwear Jersey&$10$&$13$&$54$&$8$&$Y$\\\hline
        Modern Occasion&$10$&$16$&$60$&$10$&$Y$\\\hline
        Jackets Outerwear&$7$&$9$&$9$&$7$&$Y$\\\hline
        Woven Tops&$7$&$10$&$9$&$6$&$Y$\\\hline
        Knitwear Core&$6$&$11$&$31$&$4$&$Y$\\\hline
        Knitwear Fashion&$12$&$21$&$70$&$7$&$Y$\\\hline
        Woven Tops Core&$6$&$13$&$35$&$0$&$Y$\\\hline
        Woven Tops Fashion&$13$&$24$&$74$&$4$&$Y$\\\hline 
   \hline
	\end{tabular}
	\caption{Datasets used in the experiments. For each PDF file, we extracted all the pages. $\#P$ represents number of pages, $\#T$ represents number of text blocks, $\#I$ represents number of total images present in pdf file. Here $\#H$ represents hashtags available in pdf file and $GT$ is ground truth attributes are available for the pdf file.}
\label{tbl:realdataset} 
\end{table}
\subsection{Evaluation Measures}
We use Accuracy, True Positive Rate (Recall) and  F1 score as the evaluation metrics. We compute Accuracy (denoted as $P$) as percentage of correct value generated by our framework; True positive rate (denoted as $TPR$) as percentage of ground truth value retrieved by our framework; F1 score (denoted as $F1$) as harmonic mean of Precision and Recall. 
\subsection{Baselines}
To evaluate our proposed framework, we choose the following models as baselines for text: topic rank \cite{bougouin2013topicrank} and sOpenTag \cite{xu2019scaling}. Our attribute value extraction task for images is highly related to the visual question answering tasks. Thus, we used two baselines,   Vilt \cite{kim2021vilt}, and  BLIP \cite{li2022blip} for visual attribute value extraction.
\subsection{Accuracy of \newmethod}
For all datasets we compute F1-score (\%) for text and images.  The results for qualitative measure for data is shown in Table \ref{tbl:resultsall}. We observed that F1-score (Image) is perfect for all the dataset as images provide clear visual attributes for future trends. However, text data has missing attributes and \newmethod is able to extract average $92.5\%$ of attributes from the all the PDF files.
\begin{table}[t]
	\centering
	\footnotesize
	\begin{tabular}{|c||c|c|}
	\cline{1-3}
      Dataset	& F1-score (Text) &  F1-score (Image) \\	\hline
	Boys Apparel    &$94.3\%$&$100\%$\\\hline
         Women's Cut Sew&$88.6\%$&$100\%$\\\hline
         Women's Woven Tops&$100\%$&$100\%$\\\hline
        Country Life Boys &$89.4\%$&$100\%$\\\hline
        Knitwear Jersey&$82.4\%$&$100\%$\\\hline
        Modern Occasion&$92.8\%$&$100\%$\\\hline
        Jackets Outerwear&$89.4\%$&$100\%$\\\hline
        Woven Tops&$86.6\%$&$100\%$\\\hline
        Knitwear Core&$92.5\%$&$100\%$\\\hline
        Knitwear Fashion&$97.8\%$&$100\%$\\\hline
        Woven Tops Core&$96.8\%$&$100\%$\\\hline
        Woven Tops Fashion&$98.9\%$&$100\%$\\\hline
   \hline
	\end{tabular}
	\caption{Test accuracy for multiple datasets for \newmethod}
\label{tbl:resultsall} 
\end{table}

Further we compare the performance of \newmethod with the aforementioned baseline values for text and images separately. We use "Woven Tops Core" dataset with 8 pages. We evaluate text extraction on 6 pages out of 8 with 7 attributes namely Sleeve Style, Features, Product Type, Material, Neck Style, Product Categories and Color. From these 7 attributes, we have total 33 attribute values and 9 not mentioned values. Table \ref{tbl:resultstext} shows the performance of our method compared to baseline methods. Topic rank and sOpenTag performed very well on 'Woven Tops Core' dataset. However, \newmethod outperforms baselines in-terms of F1 score and accuracy. We also note that our proposed method is $2.7\times$ faster as compared to Topic rank and sOpenTag.
\begin{table}[t]
	\centering
	\footnotesize
	\begin{tabular}{|c||c|c|c|}
	\cline{1-4}
    Dataset	& {\bf \newmethod} & {\bf Topic Rank} \cite{bougouin2013topicrank}& {\bf sOpenTag}\cite{xu2019scaling}  \\	\hline
	Precision&$100\%$&$93.3\%$&$61.4\%$\\\hline
        True Positive Rate& $93.9\%$&$42.4\%$&$73.4\%$ \\\hline
        Accuracy&$95.3\%$&$54.7\%$&$86.2\%$\\\hline
        F1-Score&$96.8\%$&$59.5\%$&$66.7\%$ \\\hline
   \hline
	\end{tabular}
	\caption{Text Attribute extraction accuracy for 'Woven Tops Core'  datasets for \newmethod and state-of-art-methods.}
\label{tbl:resultstext} 
\end{table}

We evaluate image attribute extraction on all 8 attributes namely Sleeve Style, Product Type, Material , Neck, Categories, Age, Features and Color for all 8 pages with image information. From these 8 attributes, we have total 64 visual attribute values. Table \ref{tbl:resultsimage} shows that \newmethod outperforms the baseline methods. Both the baselines were limited to pass single attribute in prompt, therefore results in consuming more time in producing the final attribute values per page. This answer our question \textbf{Q1}.
\begin{table}[t]
	\centering
	\footnotesize
	
	\begin{tabular}{|c||c|c|c|}
	\cline{1-4}
    Attributes	& {\bf \newmethod} & {\bf Vilt} \cite{kim2021vilt}& {\bf BLIP}\cite{li2022blip}  \\	\hline
	Color       &$100\%$&$87.5\%$&$87.5\%$\\\hline
    Sleeve Style    &$100\%$&$00.0\%$&$50.0\%$ \\\hline
        Product Type&$100\%$&$62.5\%$&$62.5\%$\\\hline
        Material    &$100\%$&$75.0\%$&$75.0\%$ \\\hline
        Features    &$100\%$&$50.0\%$&$25.0\%$ \\\hline
        Categories  &$100\%$&$75.0\%$&$100\%$ \\\hline
        Age Group   &$100\%$&$75.0\%$&$00.0\%$ \\\hline
        Neck        &$100\%$&$12.5\%$&$62.7\%$ \\\hline       
   \hline
	\end{tabular}
	\caption{Accuracy per attribute for 'Woven Tops Core' dataset for \newmethod and state-of-art-methods on images.}
\label{tbl:resultsimage} 
\end{table}


\subsection{Parameter of sensitivity for \newmethod}

\subsubsection{Sensitivity to LLM Prompt for Text data}
Large Language Models (LLMs) have the ability to learn new tasks on the fly, without requiring any explicit training or parameter updates. This mode of using LLMs is called in-context learning. It relies on providing the model with a suitable input prompt that contains instructions and/or examples of the desired task. Therefore, we evaluate our proposed method \newmethod for multiple prompts for text and image attribute extraction. Here, we present prompt analysis for attribute extraction from text data only. We keep the temperature parameter constant for this experiment.
\begin{itemize}
\item \textit{Prompt 1}: "Give me all clothing characteristics of a product from the following text:"
\item \textit{Prompt 2}: "Give me color, sleeve style, product type, material, cloth features, categories, and neck attributes from the following text:"
\item \textit{Prompt 3}: "I want you to act as a product attribute extractor in retail space.Given the unstructured text data, you need to find different product attributes in the text. For example: For Input as ‘Long contrast fabric Sleeve red cotton adult polo shirts for men with contemporary design element’, the attribute extractor will return color attribute is red, sleeve attribute is Long, style sleeve attribute is contrast fabric, product type attribute is polo shirts, material attribute is cotton, feature attribute is contemporary,categories is polo shirts,  gender attribute is men and neck attribute is NA. Give me attributes like color, sleeve style, product type, material, features, categories, and neck attributes from the following text: "
\end{itemize}
\begin{table}[t]
	\centering
	\footnotesize
	\begin{tabular}{|c||c|c|c|}
	\cline{1-4}
    Dataset	& {\bf Prompt 1} & {\bf Prompt 2}& {\bf Prompt 3}  \\	\hline
	Precision&$83.3\%$&$100\%$&$90.4\%$\\\hline
        True Positive Rate&$15.5\%$&$93.9\%$&$57.6\%$ \\\hline
        Accuracy&$30.9\%$&$95.3\%$&$61.9\%$\\\hline
        F1-Score&$25.6\%$&$96.8\%$&$70.3\%$\\\hline
   \hline
	\end{tabular}
	\caption{Sensitivity to LLM Prompt for 'Woven Tops Core' dataset.}
\label{tbl:resultsprompt} 
\end{table}
\begin{figure}
	\begin{center}
	    \includegraphics[clip,trim=0cm 0cm 0cm 0cm,width=0.45\textwidth]{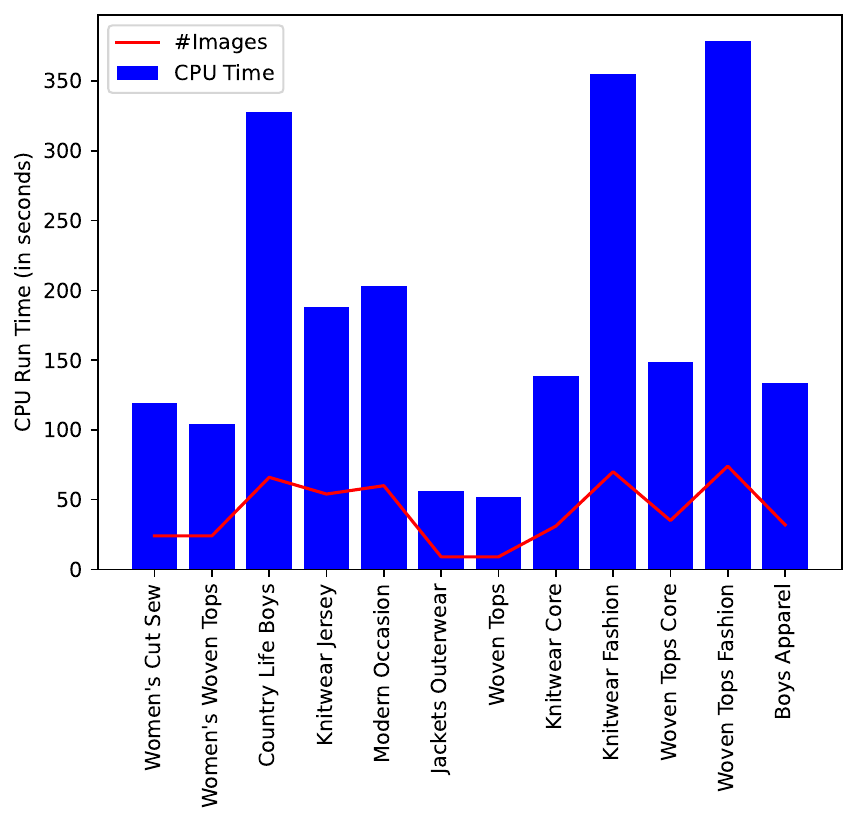}
		\caption{Running time for each PDF report.} 
		\label{fig:cputimealpdfs}
	\end{center}
\end{figure}

The table \ref{tbl:resultsprompt} shows that \textit{Prompt 2} is more effective and efficient way to extract given attributes from the text data. Too vague (\textit{Prompt 1}) and too much (\textit{Prompt 3}) information/context confuses the LLM model and therefore performance degrades. 
\subsubsection{Sensitivity to LLM Parameters}
In this study, we evaluate the performance of temperature parameter of LLM models. Temperature is a parameter in large language models (LLMs) that controls the randomness of the model's responses, ranging from 0 to 1. A higher temperature means more creative and diverse output, while a lower temperature means more predictable output. As we need predictable output, we kept the temperature parameter below 0.5. The table \ref{tbl:results4} shows that at temperature = $0.2$, method performance is high. Therefore, we chose to keep $0.2$ as parameter value.
\begin{table}[t]
	\centering
	\footnotesize
	\begin{tabular}{|c||c|c|c|c|}
	\cline{1-5}
    
	Dataset	&  {\bf $0.05$}& {\bf $0.1$} & {\bf $0.2$ } & {\bf $0.4$}\\	\hline
	Precision     &$85.7\%$&$93.7\%$&$100\%$&$100\%$\\\hline
        True Positive &$72.2\%$&$90.9\%$&$93.9\%$&$90.9\%$ \\\hline
        Accuracy      &$78.5\%$&$88.1\%$&$95.3\%$&$92.8\%$\\\hline
        F1-Score      &$84.2\%$&$92.3\%$&$96.8\%$&$95.2\%$ \\\hline
   \hline
	\end{tabular}
	\caption{Sensitivity to LLM temperature parameters for attribute extraction from text data in 'Woven Tops Core' dataset.}
\label{tbl:results4} 
\end{table}
In summary, as expected, \newmethod is sensitive to both LLM prompt and temperature parameter. This answer our question \textbf{Q2}.
\subsection{CPU Time Analysis}
In this work, we present two experiment results. First, we provide CPU time for each dataset in figure \ref{fig:cputimealpdfs}. We observe that CPU time is directly proportional to number of images in the PDF files. "Women Tops Fashion" PDF file has 74 images and took around $350$ seconds to get attributes for all $13$ pages. 
\begin{figure}
	\begin{center}
	    \includegraphics[clip,trim=0cm 0cm 0cm 0cm,width=0.45\textwidth]{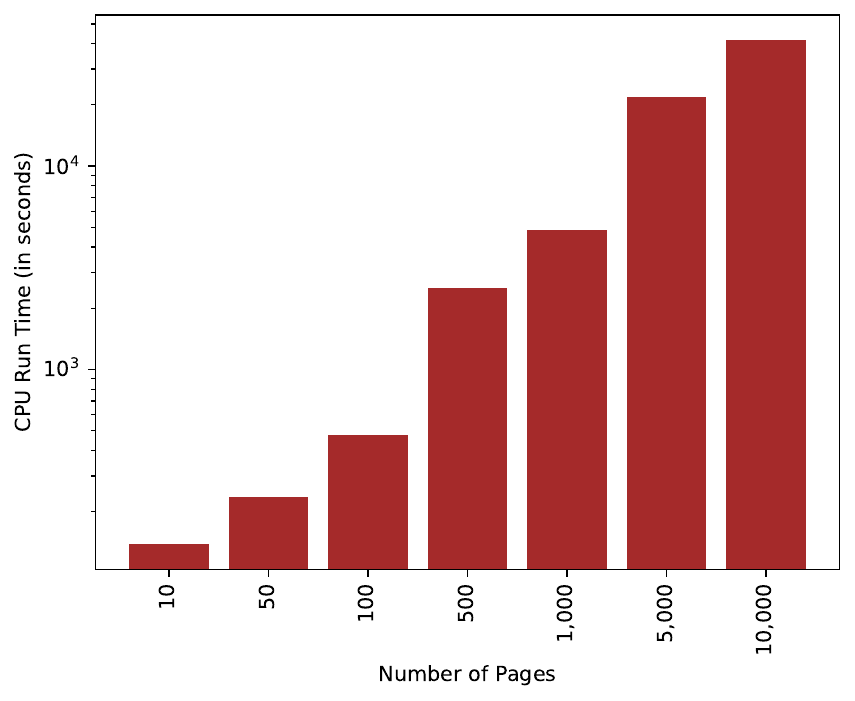}
		\caption{Running time for synthetic PDF report.} 
		\label{fig:cputimealpdfs_1}
	\end{center}
\end{figure}
Second, we created synthetic PDF files with each consisting of 500-1000 words and 4-6 images per page. We created PDF files with $[10,50,100,500,1000,5000,10000]$ pages. The figure \ref{fig:cputimealpdfs} shows that CPU time is linear w.r.t size of the PDF file. This answer our question \textbf{Q3}.

\section{CONCLUSIONS AND FUTURE WORK}
\label{sec:conclusions}
In this work, we have described an attribute extraction framework namely \newmethod for applications in the domain of inventory/E-commerce business. The goal is given a PDF file consisting of upcoming trends in the form of text and images, our framework correctly extracts the defined attributes in order to plan the future assortment and associate these predicted attributes to the current product in a catalog for better planning. The performance of the proposed method is assessed via experiments on real-world datasets.   We summarize our contribution as:
\begin{itemize}
	\item  The proposed framework effectively identifies the attributes of from PDF files to achieve an assortment planning task. To further enhance the capability, we tailored our proposed framework towards flexibility where extraction of data and attributes can be easily enhanced and modified for domain-specific applications.
	\item Through experimental evaluation on multiple datasets, we show that \newmethod  provides accurate attributes and is significantly faster in terms of CPU run time.
\end{itemize}
There is still room for improving our methods. One direction is to explore LLM models which can take sets of images \& text and provide consolidate attributes. Another direction is to further improve the product matching system that consists of product images so that our method can be more suitable for customers during the search for different products on e-commerce websites.
 
\section{Acknowledgements}
{
Research was supported by the Walmart. Any opinions, findings, and conclusions or recommendations expressed in this material are those of the author(s) and do not necessarily reflect the views of the funding parties.
}
\balance
\bibliographystyle{plain}
\bibliography{BIB/refs}
\end{document}